\documentclass[11pt,a4paper]{article}

\usepackage[utf8]{inputenc}
\usepackage[T1]{fontenc}
\usepackage{times}
\usepackage[margin=2.5cm]{geometry}
\usepackage{amsmath,amssymb}
\usepackage{graphicx}
\usepackage{booktabs}
\usepackage{hyperref}
\usepackage{array}
\usepackage{xcolor}
\usepackage{caption}
\usepackage{enumitem}

\hypersetup{
    colorlinks=true,
    linkcolor=blue!70!black,
    citecolor=green!50!black,
    urlcolor=blue!70!black,
}

\title{\textbf{Ternary Mamba: Grouped Quantization-Aware Training\\of W1.58A16 State Space Models}}

\author{
    Ramprasath Ganesaraja \\
    EdgeVerve Systems Limited \\
    \texttt{ramprasath.g@edgeverve.com}
    \and
    Sahil Dilip Panse \\
    EdgeVerve Systems Limited \\
    \texttt{SahilDilip\_Panse@edgeverve.com}
    \and
    Swathika N \\
    EdgeVerve Systems Limited \\
    \texttt{swathika.n@edgeverve.com}
}

\date{}

\begin{document}
\maketitle

\begin{abstract}
State Space Models (SSMs) such as Mamba-2~\cite{mamba2} offer linear-time inference but their memory footprint limits edge deployment. Prior ternary SSM work (Slender-Mamba) trains from scratch on 150B tokens; we show a pretrained checkpoint suffices, reducing the marginal token budget by 1,000$\times$. Using \textbf{grouped quantization-aware training (QAT) with knowledge distillation} from a frozen FP16 teacher, we compress Mamba-2 1.3B to \textbf{3.61$\times$} (2,687$\to$744\,MB) and achieve \textbf{48.1\% zero-shot accuracy} (7-task average) in just \textbf{102M tokens} (4 GPU-hours, single H100)---approaching Bi-Mamba's 48.4\% (within $\pm$0.9pp CI). This QAT-from-pretrained setting reveals \emph{zero-ratio collapse}, a novel instability caused by learnable quantization scales that does not arise in from-scratch training. We further show that post-hoc correction strategies effective for Transformers fail for SSMs due to error accumulation through the recurrence. These results demonstrate that ternary SSMs do not require expensive from-scratch training: QAT from pretrained checkpoints with KD is a data-efficient alternative.
\end{abstract}

\section{Introduction}

Large language models face a fundamental tension between capability and deployability. Two orthogonal approaches address this: \emph{architectural efficiency} (Mamba-2's $O(L)$ selective SSM replacing $O(L^2)$ attention) and \emph{weight compression} (BitNet b1.58's ternary $\{-1, 0, +1\}$ weights achieving near-parity with FP16 at 3B+ parameters).

Their intersection remains underexplored. Slender-Mamba~\cite{slendermamba} applies BitNet-style ternary quantization to Mamba-2 170M but trains from scratch on 150B tokens with per-tensor scaling. Other Mamba quantization---Quamba~\cite{quamba} (W4A8 PTQ), MambaQuant~\cite{mambaquant} (8-bit PTQ), Bi-Mamba~\cite{bimamba} (1-bit from scratch)---is either post-training at $\geq$4 bits or binary with per-column scaling trained on 105B tokens. PTQ below 4 bits degrades catastrophically for SSMs~\cite{quamba,mambaptq}; we confirm this extends to the ternary regime (\S\ref{sec:main_results}), with naive W1.58 PTQ producing PPL $\sim$13M (effectively random). The key open question is whether ternary SSMs can be obtained cheaply via QAT from a pretrained checkpoint, avoiding the enormous from-scratch cost.

The SSM recurrence $h_t = \bar{A}_t h_{t-1} + \bar{B}_t x_t$ presents a distinct quantization challenge: errors accumulate as $e_T = \sum_{t=1}^{T} \bar{A}^{T-t} \epsilon_t$, mixing through the state transition rather than appearing additively as in Transformer attention. This makes post-hoc correction ineffective and necessitates quantization-aware training.

\paragraph{Contributions:}
\begin{enumerate}[leftmargin=*]
    \item \textbf{Grouped ternary QAT for pretrained SSMs:} To our knowledge, the first application of per-group ternary QAT to a pretrained SSM---Mamba-2 1.3B at W1.58A16, 3.61$\times$ compression, approaching Bi-Mamba accuracy (48.1\% vs 48.4\%) with substantially lower marginal cost (102M tokens, 4 GPU-hours).
    \item \textbf{Zero-ratio collapse:} Discovery of a novel ternary SSM instability caused by learnable quantization scales, resolved by non-learnable absmean recomputation.
    \item \textbf{SSM quantization characterization:} Systematic evidence that post-hoc correction (Kalman, James-Stein) and noise-shaping (sigma-delta) do not transfer from Transformers to SSMs.
    \item \textbf{Scaling and ablations:} Group-size optimality at $g=128$, layer-selective Pareto frontier, and log-linear scaling projecting continued improvement.
\end{enumerate}

\paragraph{Scope caveat:} This work focuses exclusively on Mamba-2 1.3B, a single SSM architecture trained on English (C4 corpus). Generalization to other SSM variants (Mamba-1, Mamba-3, Jamba), larger model scales (2.7B+), and non-English languages remains unexplored and should be studied before claiming universal applicability of the proposed ternary QAT approach.

\section{Related Work}

\paragraph{Mamba quantization.} Quamba~\cite{quamba} and Quamba2~\cite{quamba2} apply W4A8/W8A8 PTQ with smooth quantization. MambaQuant~\cite{mambaquant} uses KLT-enhanced rotation for 8-bit PTQ. Mamba-PTQ~\cite{mambaptq} identifies scattered outliers in SSMs that make standard quantization harder. None achieve below 4-bit without catastrophic degradation.

\paragraph{Binary and ternary SSMs.} Bi-Mamba~\cite{bimamba} trains binary $\{-1, +1\}$ Mamba-2 from scratch on 105B tokens with per-column scales and bias, achieving 48.4\% zero-shot average at 1.3B scale. Slender-Mamba~\cite{slendermamba} applies BitNet-style ternary quantization to Mamba-2 170M, training from scratch on 150B tokens with per-tensor scaling. Our approach differs from both: (1) per-group ($g=128$) scaling rather than per-tensor or per-column, (2) we start from a pretrained checkpoint requiring only 102M QAT tokens, (3) 1.3B scale (7.6$\times$ larger than Slender-Mamba), and (4) we do not require from-scratch training.

\paragraph{Ternary Transformers.} BitNet b1.58~\cite{bitnet} achieves ternary Transformer parity with FP16 at 3B+ scale from scratch on 100B tokens. ParetoQ~\cite{paretoq} establishes ternary scaling laws. To our knowledge, our work is the first to apply grouped ternary QAT to a pretrained SSM at the 1.3B scale.

\paragraph{Post-hoc error correction.} LREC~\cite{lrec} and QEP~\cite{qep} apply linear corrections to quantized Transformer outputs. We test analogous corrections for SSMs and show they fail---a fundamental architectural distinction.

\section{Method}

\subsection{TernaryLinear Module}

Each \texttt{in\_proj} and \texttt{out\_proj} (96 modules, 85.7\% of parameters) is replaced with a TernaryLinear that groups the weight matrix into blocks of $g=128$ elements, computes a non-learnable per-group scale $s_g = \frac{1}{g}\sum_{i \in g}|w_i|$, ternarizes via $\tilde{w}_i = \text{round}(\text{clip}(w_i / s_g, -1, 1))$, and dequantizes as $\hat{w}_i = s_g \cdot \tilde{w}_i$. Gradients flow through the straight-through estimator (STE)~\cite{ste}.

The scale is recomputed each forward pass (not a learned parameter). This design choice prevents zero-ratio collapse (\S\ref{sec:collapse}): the self-regulating property of absmean creates a negative feedback loop that maintains stable $\sim$26\% sparsity.

\subsection{Quantization Coverage}

Quantized: \texttt{in\_proj} and \texttt{out\_proj} in all 48 blocks (85.7\% of parameters). Kept in FP16: SSM dynamics ($A_{\log}$, $D$, $dt\_bias$---$<$0.3\% of parameters), conv1d, embedding, lm\_head, and norms. This follows Quamba~\cite{quamba} and Bi-Mamba~\cite{bimamba}, which report degradation from quantizing SSM parameters below 8 bits.

\subsection{Training Objective}

We train from the pretrained FP16 Mamba-2 1.3B checkpoint with a frozen FP16 teacher providing knowledge distillation:
\begin{equation}
\mathcal{L} = \alpha \cdot D_\text{KL}(p_\text{teacher} \| p_\text{student}) \cdot T^2 + (1-\alpha) \cdot \mathcal{L}_\text{CE}
\end{equation}
with $\alpha = 0.5$, $T = 1.0$. Optimizer: AdamW ($\beta=(0.9, 0.95)$, decay 0.01), lr $2.5\times10^{-4}$ with cosine schedule over 50k steps (1k warmup). Data: C4~\cite{c4} (English), batch $2 \times$ seq $1024 = 102$M total tokens. Hardware: single NVIDIA H100, $\sim$4 hours.

\section{Zero-Ratio Collapse}
\label{sec:collapse}

Training with a learnable scale parameter ($s_g$ as \texttt{nn.Parameter}) causes catastrophic sparsity growth:

\begin{table}[h]
\centering
\caption{Zero-ratio collapse under different scale parameterizations.}
\label{tab:collapse}
\begin{tabular}{lcccc}
\toprule
Experiment & Steps & Scale type & Zero ratio & WikiText-2 PPL \\
\midrule
exp001 & 5k & Fixed absmean & 57.6\% & 22.54 \\
exp002 & 50k & Learnable & \textbf{90.3\%} & \textbf{47.45} \\
exp006 & 50k & Fixed absmean & 26.1\% & 11.25 \\
\bottomrule
\end{tabular}
\end{table}

The mechanism is a positive feedback loop: the optimizer increases $s_g$ to reduce short-term KD loss $\to$ wider zero band $\to$ more weights collapse to zero $\to$ capacity loss $\to$ further scale increase. The fixed absmean formulation breaks this loop: if zero ratio increases, absmean decreases, the threshold shrinks, and fewer weights round to zero---a self-regulating equilibrium at $\sim$26\%.

\begin{figure}[h]
\centering
\includegraphics[width=0.7\textwidth]{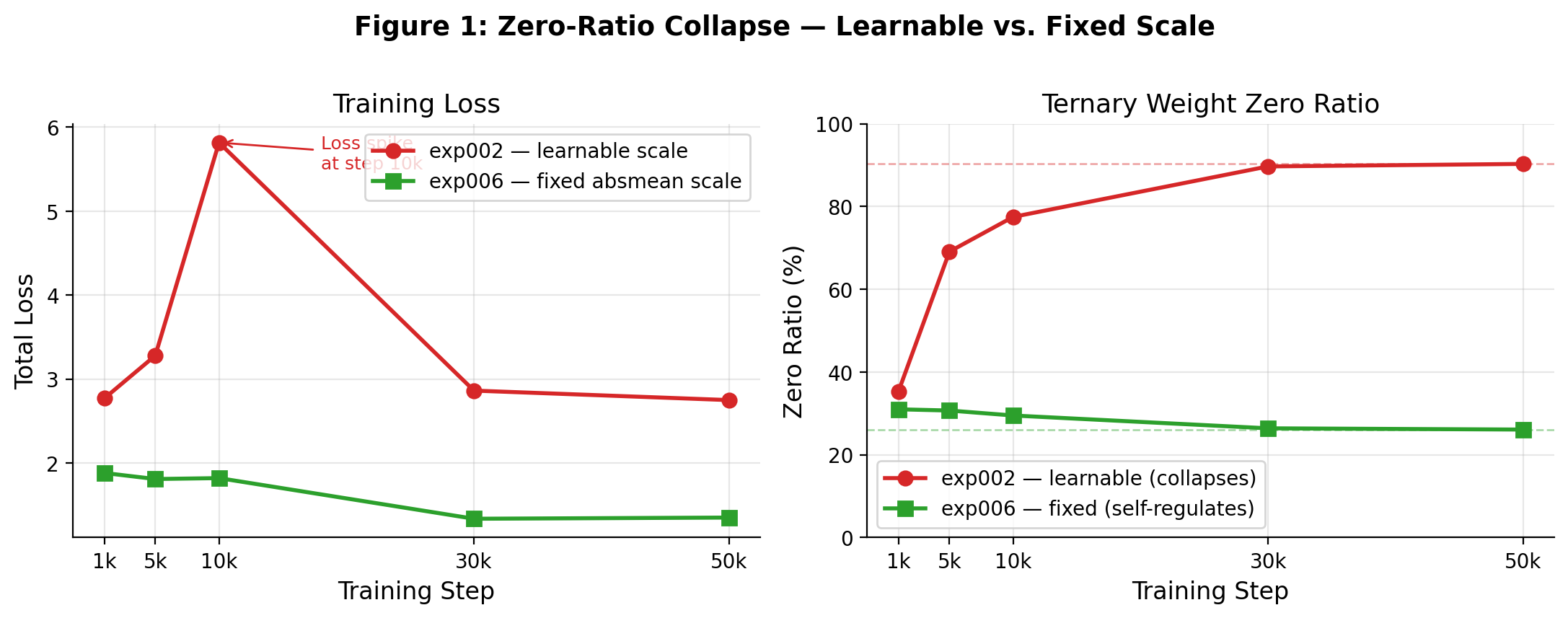}
\caption{Learnable scale (exp002) collapses to 90\% zeros with a loss spike at 10k steps, while fixed absmean (exp006) self-regulates at $\sim$26\%.}
\label{fig:collapse}
\end{figure}

\section{Experiments}

\subsection{Main Results}
\label{sec:main_results}

\begin{table}[h]
\centering
\caption{Primary results. All models evaluated on WikiText-2 test (285K tokens) and C4 held-out (512K tokens).}
\label{tab:main}
\begin{tabular}{lccccc}
\toprule
Model & Bits & Compr. & WT2 PPL & C4 PPL & 7-task avg \\
\midrule
Mamba-2 1.3B FP16 (teacher) & 16 & 1.00$\times$ & 11.69$\pm$0.05 & 17.47$\pm$0.05 & 55.5\% \\
Ternary PTQ g128 (no training) & 1.58 & 3.61$\times$ & $\sim$13.2M & $\sim$23.0M & N/A \\
\textbf{Ternary Mamba g128 (ours)} & \textbf{1.58} & \textbf{3.61$\times$} & \textbf{20.31$\pm$0.11} & \textbf{21.67$\pm$0.05} & \textbf{48.1\%} \\
Ternary Mamba, layer-sel N=2 & $\sim$2.1 & 2.91$\times$ & 19.63$\pm$0.11 & 20.99$\pm$0.05 & 46.4\% \\
\bottomrule
\end{tabular}
\end{table}

The domain-matched C4 gap is +4.20 PPL (+24\%) vs.\ FP16. The WikiText-2 gap (+8.62) is inflated by train/eval domain mismatch (model trained on C4). Multi-seed replication (seeds 42, 137, 2024) confirms tight variance: C4 PPL std = 0.01, 7-task std = $\pm$0.13pp. Crucially, naive ternary PTQ (same quantizer, zero training) produces PPL $\sim$13M---effectively random output---establishing that \textbf{QAT is necessary, not merely beneficial}, for ternary SSMs.

\subsection{Comparison with Prior Work}

\begin{table}[h]
\centering
\caption{Zero-shot accuracy comparison (7-task suite).}
\label{tab:comparison}
\resizebox{\textwidth}{!}{%
\begin{tabular}{lccccccccccc}
\toprule
Method & Bits & Type & Train tok. & BoolQ & PIQA & Hella. & WinoG & ARC-E & ARC-C & OBQA & Avg \\
\midrule
Mamba-2 FP16 & 16 & --- & --- & 62.9 & 73.7 & 59.9 & 61.1 & 60.4 & 33.2 & 37.4 & 55.5 \\
Bi-Mamba 1.3B~\cite{bimamba} & 1 & QAT scratch & 105B & 60.0 & 68.8 & 47.3 & 55.9 & 48.0 & 26.3 & 32.2 & 48.4 \\
\textbf{Ternary Mamba (ours)} & \textbf{1.58} & \textbf{QAT} & \textbf{102M} & \textbf{56.0} & \textbf{69.3} & \textbf{47.5} & \textbf{56.3} & \textbf{48.2} & \textbf{25.9} & \textbf{33.8} & \textbf{48.1} \\
BitNet b1.58 1.3B~\cite{bitnet}$^\dagger$ & 1.58 & QAT scratch & 100B & 56.7 & 68.8 & 37.7 & 55.8 & 54.9 & 24.2 & 19.6 & 45.4 \\
\bottomrule
\end{tabular}%
}
\smallskip

{\footnotesize $^\dagger$BitNet is a Transformer (LLaMA architecture), included for cross-architecture context. Per-task 95\% CIs from Wilson score intervals; 7-task average CI $\approx \pm$0.9pp. The 0.3pp gap between our method and Bi-Mamba is within this CI.}
\end{table}

\subsection{Training Efficiency}

The marginal QAT cost is 4 GPU-hours on a single H100, compared to Bi-Mamba's 5,780 GPU-hours (32$\times$ A100). Note that this comparison is favorable because it excludes the amortized FP16 pretraining cost ($\sim$4,000 GPU-hours), which we inherit from the publicly available checkpoint. The marginal cost nonetheless makes the approach accessible to single-GPU researchers.

\begin{table}[h]
\centering
\caption{Training cost comparison.}
\label{tab:efficiency}
\begin{tabular}{lcccc}
\toprule
Method & Pretraining & Quantization & Total & 7-task avg \\
\midrule
Bi-Mamba~\cite{bimamba} (from scratch) & 0 & 5,780 GPU-hrs & 5,780 GPU-hrs & 48.4\% \\
\textbf{Ternary Mamba (ours)} & \textbf{0$^\ddagger$} & \textbf{4 GPU-hrs} & \textbf{4 GPU-hrs} & \textbf{48.1\%} \\
\bottomrule
\end{tabular}

\smallskip
{\footnotesize $^\ddagger$FP16 checkpoint publicly available on HuggingFace (\texttt{state-spaces/mamba2-1.3b}); cost amortized over all downstream uses.}
\end{table}

\subsection{Scaling Behavior}

C4 PPL improves monotonically through 307M tokens with no plateau:

\begin{table}[h]
\centering
\caption{Scaling behavior across training budget.}
\label{tab:scaling}
\begin{tabular}{lccc}
\toprule
Run & Total tokens & C4 PPL & $\Delta$ vs baseline \\
\midrule
exp007 (primary) & 102M & 21.67$\pm$0.05 & --- \\
exp015 (+20k, low-LR) & 184M & 21.12$\pm$0.05 & $-$0.55 \\
exp016 (+20k, low-LR) & 266M & 21.04$\pm$0.05 & $-$0.63 \\
exp017 (+50k, full-LR) & 307M & \textbf{20.65$\pm$0.05} & \textbf{$-$1.02} \\
\bottomrule
\end{tabular}
\end{table}

The monotonically decreasing trend is statistically significant (Mann-Kendall exact test: $S=-6$, one-sided $p=0.042$; OLS regression slope test on log-tokens: $p=0.020$, $R^2=0.92$). The overall improvement of $-$1.02 PPL is 20$\times$ the bootstrap 95\% CI ($\pm$0.05), and 3 of 4 consecutive drops exceed 2$\times$CI individually. A log-linear fit is consistent with extrapolations projecting FP16 parity in the 1--10B token range, though such extrapolation from only four data points should be interpreted with caution.

\begin{figure}[h]
\centering
\includegraphics[width=0.65\textwidth]{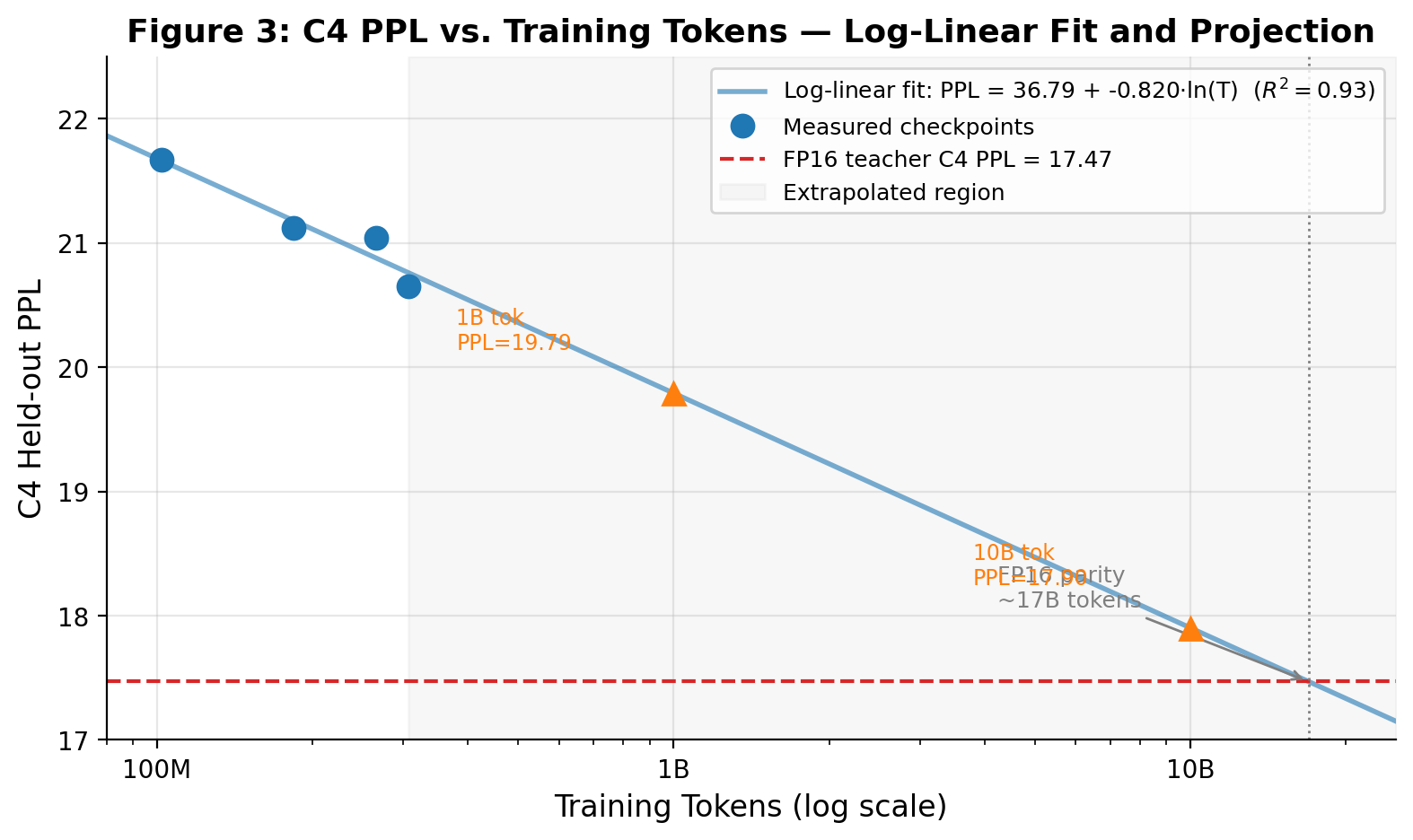}
\caption{C4 PPL vs.\ training tokens (log scale). Monotonic decrease with no plateau observed; trajectory is consistent with continued improvement under additional training.}
\label{fig:scaling}
\end{figure}

\subsection{Group-Size Ablation}

All configurations trained for 50k steps on C4:

\begin{table}[h]
\centering
\caption{Group-size ablation (50k steps, C4).}
\label{tab:groupsize}
\begin{tabular}{lccc}
\toprule
Group size & WikiText-2 PPL & C4 PPL & Top-1 agree \\
\midrule
$g=64$ & 20.28$\pm$0.11 & 21.64$\pm$0.91 & 69.9\% \\
\textbf{$g=128$} & \textbf{20.31$\pm$0.11} & \textbf{21.66$\pm$0.91} & \textbf{71.2\%} \\
$g=256$ & 20.34$\pm$0.11 & 21.68$\pm$0.92 & 67.1\% \\
\bottomrule
\end{tabular}
\end{table}

The PPL range across a 4$\times$ granularity sweep is 0.06---statistically indistinguishable. The quality bottleneck is the ternary constraint itself, not scale resolution. We select $g=128$ as optimal based on peak top-1 agreement.

\begin{figure}[h]
\centering
\includegraphics[width=0.65\textwidth]{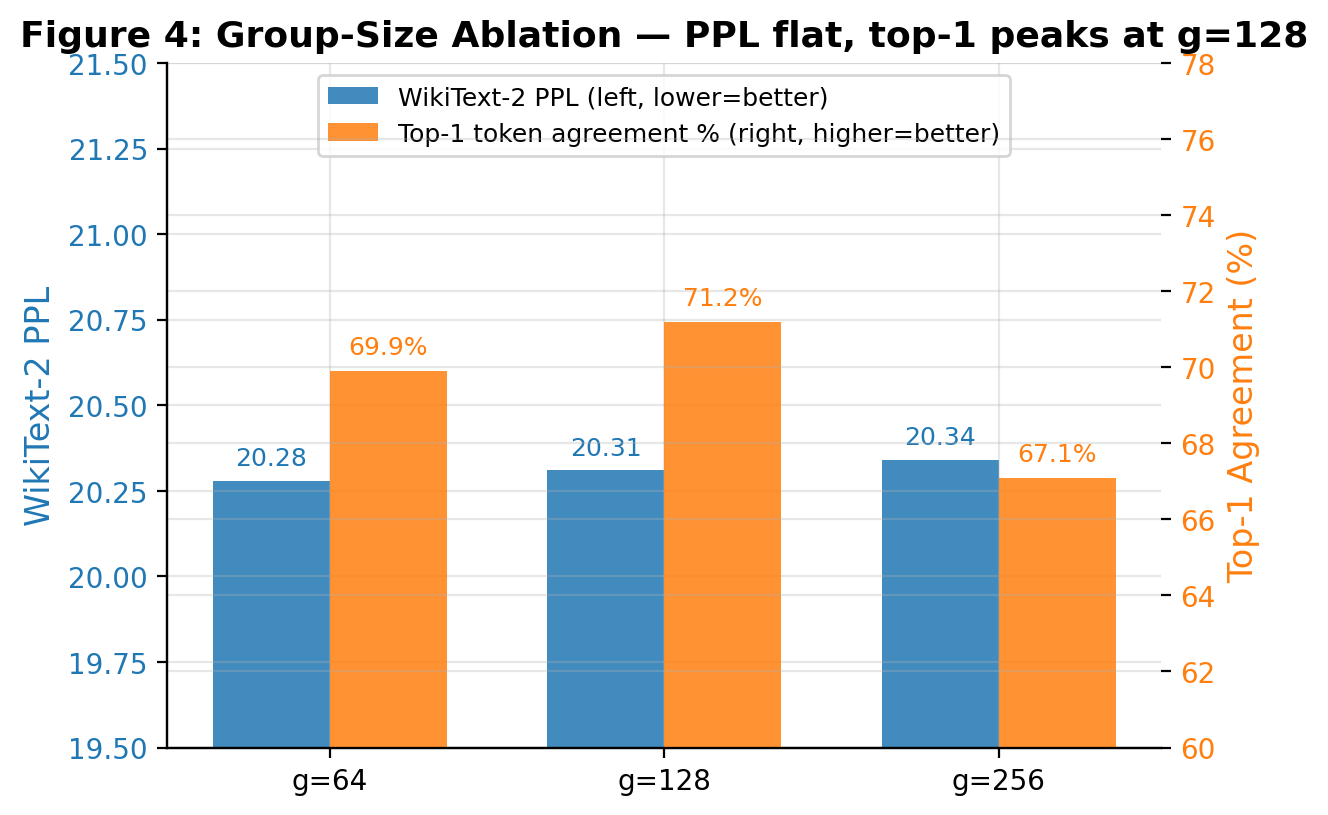}
\caption{Group-size ablation. PPL is flat across $g\in\{64, 128, 256\}$; top-1 agreement peaks at $g=128$.}
\label{fig:groupsize}
\end{figure}

\subsection{KD Mixing Weight Ablation}

\begin{table}[h]
\centering
\caption{KD mixing weight ($\alpha$) ablation.}
\label{tab:alpha}
\begin{tabular}{lcc}
\toprule
$\alpha$ & Loss components & C4 PPL \\
\midrule
0.0 & CE only & 87.55$\pm$3.86 \\
\textbf{0.5} & \textbf{KD + CE} & \textbf{21.67$\pm$0.01} \\
1.0 & KD only & 37.96$\pm$1.67 \\
\bottomrule
\end{tabular}
\end{table}

The balanced split outperforms both extremes by large margins. KD provides the dominant training signal (CE-only degrades 4$\times$) while CE anchoring prevents drift (KD-only degrades 1.75$\times$). The KD-only failure mechanism: without ground-truth labels, the student inherits and amplifies teacher errors on low-confidence tokens---the KL objective provides no corrective signal when the teacher assigns probability mass to incorrect tokens, causing the student to drift toward over-confident wrong predictions.

\section{Why Post-Hoc Correction Fails for SSMs}

\subsection{Experiments}

We tested two post-hoc correction strategies applied via forward hooks (no retraining):
\begin{itemize}[leftmargin=*]
    \item \textbf{Kalman gain:} Amplification $\hat{y} = y + Ky$, sweeping $K \in \{0.05, 0.10, 0.20, 0.30, 0.50\}$
    \item \textbf{James-Stein shrinkage~\cite{jamesstein}:} $\hat{y} = y \cdot \max(0, 1 - c/\|y\|^2)$, sweeping 12 values over 3 orders of magnitude
\end{itemize}

\begin{table}[h]
\centering
\caption{Post-hoc correction results.}
\label{tab:posthoc}
\begin{tabular}{lccc}
\toprule
Method & Best PPL & Baseline & Result \\
\midrule
Kalman gain & 24.65 (+2.61) & 22.04 & Monotonically worse \\
James-Stein & 22.06 (+0.01) & 22.04 & Monotonically worse \\
No correction & \textbf{22.04} & 22.04 & Global minimum \\
\bottomrule
\end{tabular}
\end{table}

\begin{figure}[h]
\centering
\includegraphics[width=0.65\textwidth]{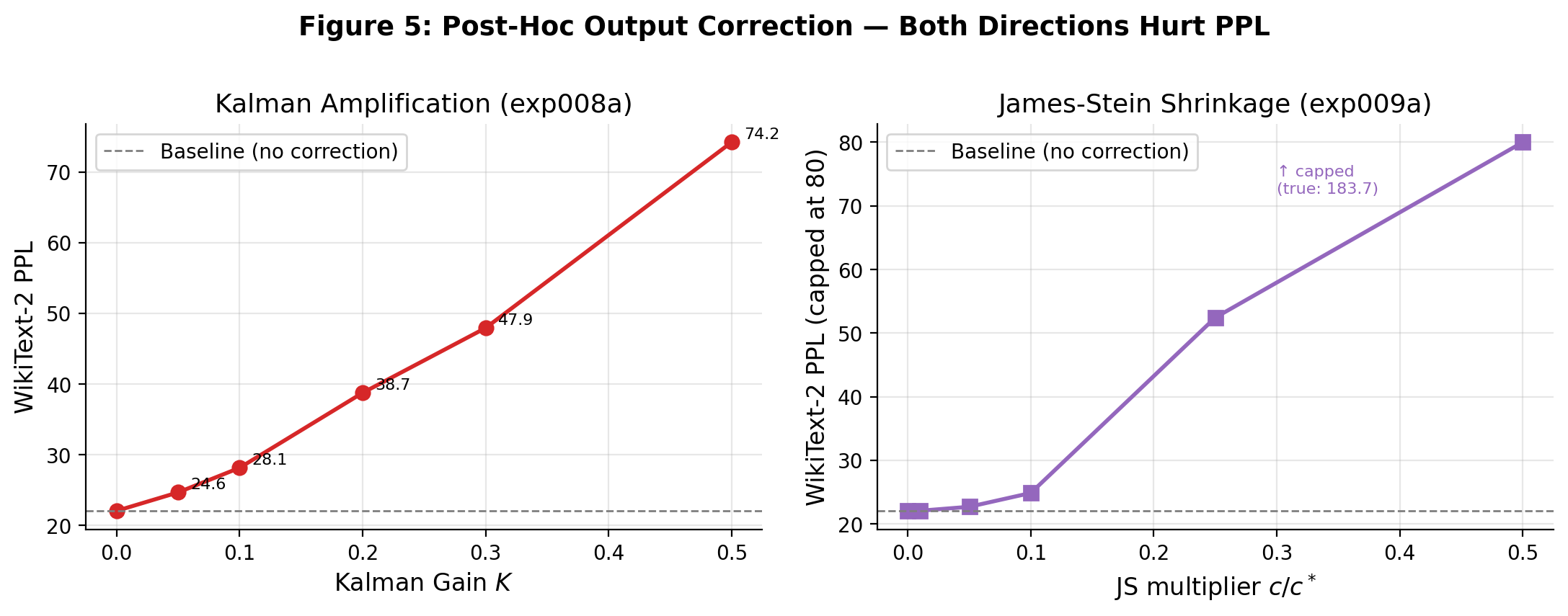}
\caption{Both Kalman amplification and James-Stein shrinkage increase PPL monotonically from baseline. The optimum is zero correction.}
\label{fig:posthoc}
\end{figure}

\subsection{Analysis}

In Transformers, attention computes a weighted sum of value vectors---quantization errors are approximately additive on the output and amenable to linear correction. In SSMs, the recurrence mixes error into the hidden state across all prior positions:
\begin{equation}
e_T = \sum_{t=1}^{T} \bar{A}^{T-t} \epsilon_t
\end{equation}

The error at position $T$ depends on the full sequence history $\{\epsilon_k\}_{k \leq t}$ filtered through the state dynamics. No pointwise function of the current output can disentangle this---formally, $\min_f \mathcal{L}(f(y)) \geq \mathcal{L}(y)$ for any scalar $f$, because the error is structural and history-dependent rather than additive and independent.

\subsection{Sigma-Delta Noise-Shaping}

We additionally tested sigma-delta ($\Sigma\Delta$) modulation---successful in 1-bit DACs for temporally correlated signals. The method requires within-group smoothness ($\text{TV}(x) < \text{L1}(x)$). Measuring the TV/L1 ratio across all 96 projection inputs in Mamba-2:

\begin{table}[h]
\centering
\caption{TV/L1 diagnostic for sigma-delta applicability.}
\label{tab:sigmadelta}
\begin{tabular}{lc}
\toprule
Signal & TV/L1 \\
\midrule
i.i.d.\ Gaussian (theoretical) & 1.414 \\
Mamba grand mean (96 hooks) & \textbf{1.421} \\
\bottomrule
\end{tabular}
\end{table}

Every layer exceeds the Gaussian baseline---Mamba's representations are more decorrelated than random noise along the feature dimension, as expected for disentangled learned representations. $\Sigma\Delta$ is inapplicable.

\begin{figure}[h]
\centering
\includegraphics[width=0.65\textwidth]{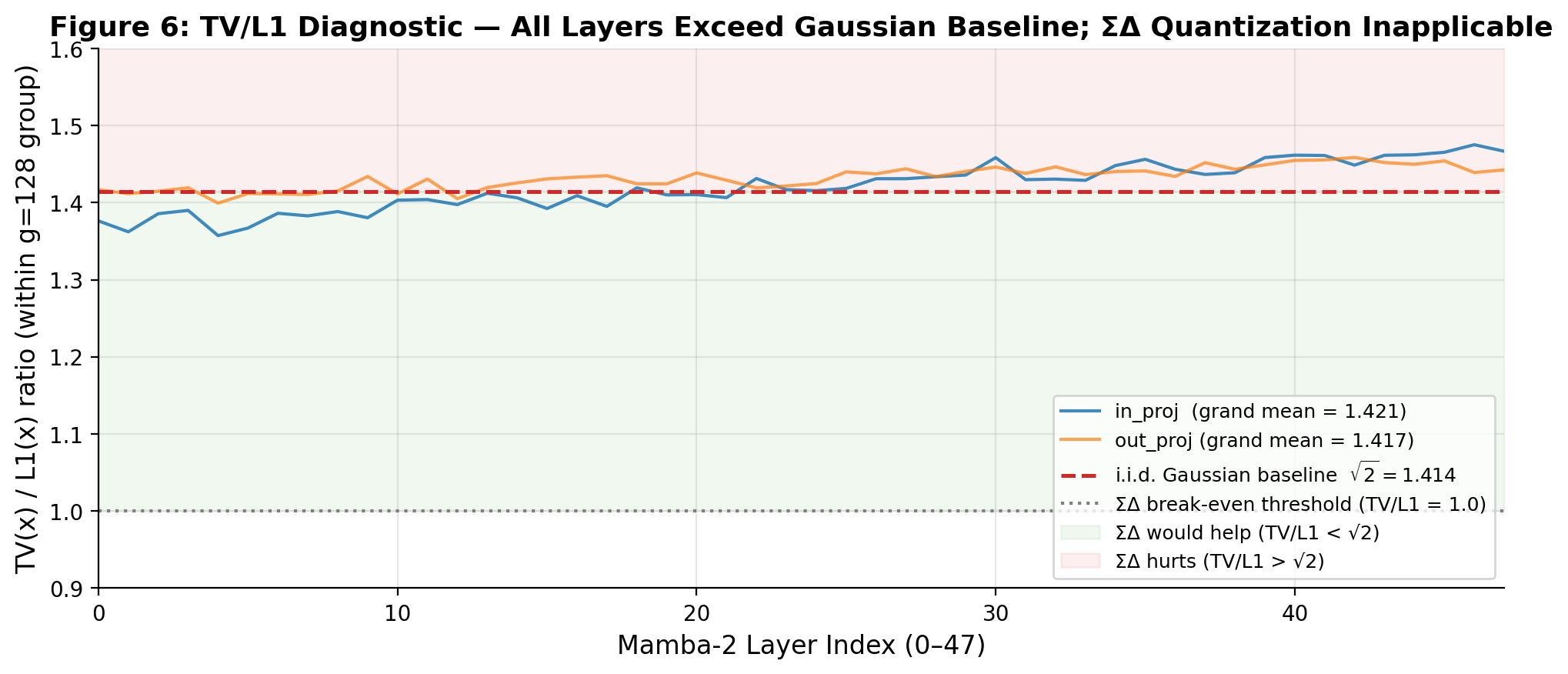}
\caption{TV/L1 ratio exceeds the Gaussian baseline ($\sqrt{2}$) at every layer, ruling out sigma-delta noise-shaping for SSMs.}
\label{fig:tvl1}
\end{figure}

\section{Layer-Selective Quantization}

Keeping the first $N$ and last $N$ blocks in FP16 (i.e., $2N$ blocks total; $N=2$ means 4 FP16 blocks: layers 0, 1, 46, 47) while quantizing the middle $48-2N$ provides a continuous compression-quality trade-off:

\begin{table}[h]
\centering
\caption{Layer-selective quantization trade-off.}
\label{tab:layersel}
\begin{tabular}{lcccc}
\toprule
Config & Ternary blocks & Compression & WikiText-2 PPL & $\Delta$ vs FP16 \\
\midrule
All ternary ($N=0$) & 48/48 & 3.61$\times$ & 20.31$\pm$0.11 & +8.62 \\
Layer-sel $N=2$ & 44/48 & 2.91$\times$ & 19.63$\pm$0.11 & +7.94 \\
FP16 ($N=24$) & 0/48 & 1.00$\times$ & 11.69$\pm$0.05 & 0 \\
\bottomrule
\end{tabular}
\end{table}

$N=2$ recovers 7.9\% of the PPL gap while protecting 8.3\% of blocks---near-linear proportionality. Direct measurement of per-layer weight quantization error confirms uniform distribution: NRMSE coefficient of variation (CV = std/mean computed over all 48 layers) is $<$0.005 for both \texttt{in\_proj} and \texttt{out\_proj} independently (mean NRMSE = 0.5699, std = 0.0012--0.0027). No single layer dominates the error.

\begin{figure}[h]
\centering
\includegraphics[width=0.65\textwidth]{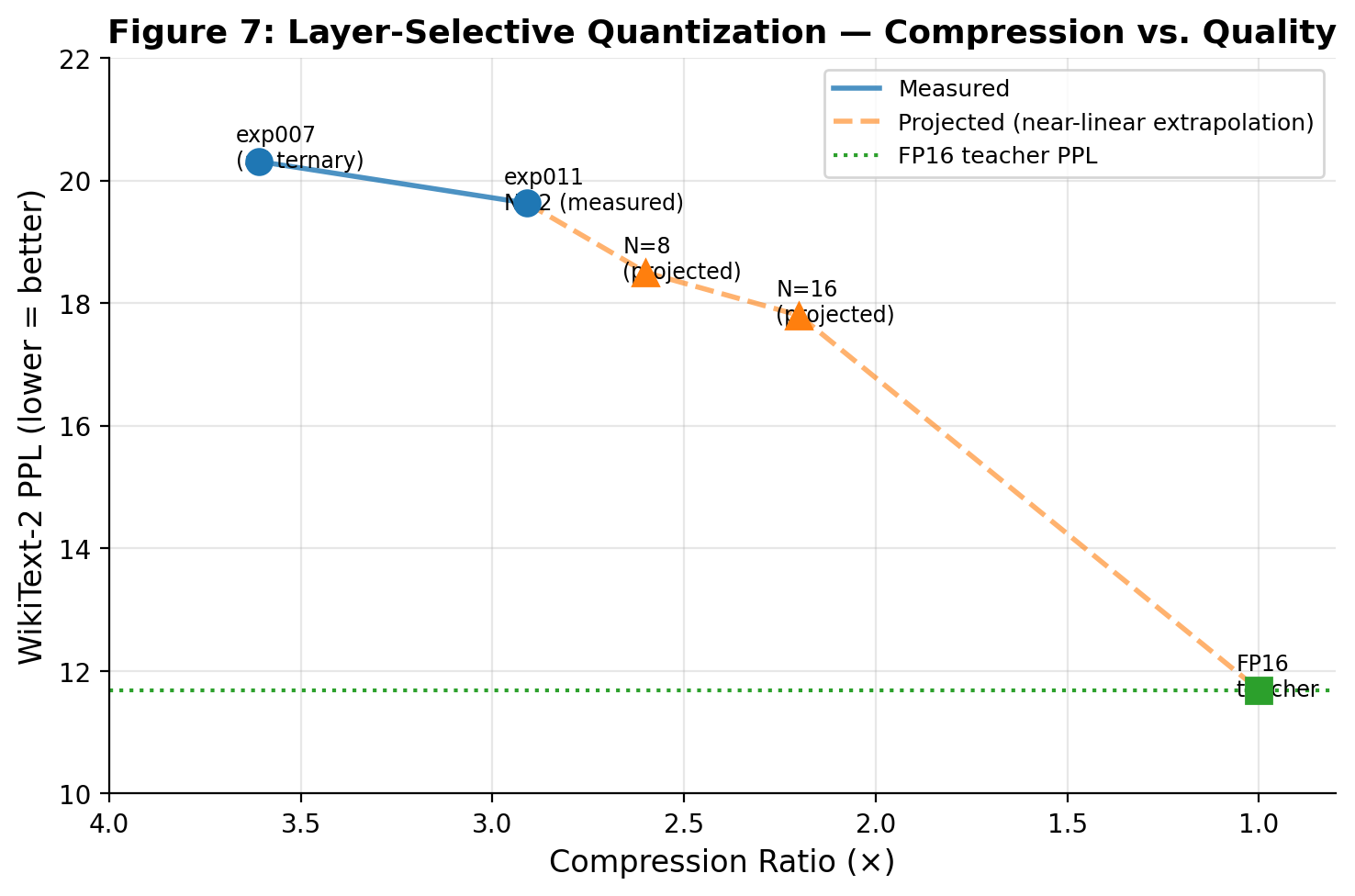}
\caption{Layer-selective quantization Pareto frontier. PPL recovery is approximately linear in the fraction of protected layers.}
\label{fig:layersel}
\end{figure}

\section{Storage Compression vs.\ Inference}

The 3.61$\times$ compression is a \emph{storage and deployment-memory} claim. The current QAT implementation (dequantized simulation) uses more GPU memory than FP16 due to latent weight storage and quantization intermediates. Realizing inference benefits requires packed 2-bit weight storage with a custom ternary GEMM kernel.

\begin{table}[h]
\centering
\caption{Measured inference memory (NVIDIA H100 NVL, seq\_len=1024).}
\label{tab:memory}
\begin{tabular}{lccc}
\toprule
Configuration & FP16 & Ternary (sim.) & Ternary (packed, proj.) \\
\midrule
Model weights & 2,599 MB & 5,322 MB & $\sim$744 MB \\
Peak memory (bs=1) & 2,932 MB & 6,932 MB & $\sim$1,100 MB$^*$ \\
\bottomrule
\end{tabular}

\smallskip
{\footnotesize $^*$Projected with packed ternary weights; activation memory remains in FP16.}
\end{table}

\paragraph{Kernel landscape.} BitBLAS (Microsoft, 2024) and tinyBLAS provide ternary/binary GEMM kernels for Transformer architectures; TernaryLLM~\cite{ternaryllm} demonstrates W1.58 inference with custom CUDA kernels achieving 5--8$\times$ GEMM speedup on weight-loading-bound layers. For Mamba-2's specific projection shapes (2048$\times$8192, 4096$\times$2048), no kernel exists yet, but the W1.58A16 regime enables the same conditional add/subtract optimization: each multiply-accumulate reduces to a sign-conditional addition, eliminating the multiplier entirely.

\paragraph{Theoretical speedup.} At batch size 1 (memory-bandwidth-bound), throughput scales inversely with weight memory traffic: 3.61$\times$ compression implies up to 3.61$\times$ decoding throughput. At larger batch sizes (compute-bound), the benefit comes from replacing FP16 multiply-accumulate with ternary add/subtract, yielding 5--8$\times$ GEMM speedup on hardware with efficient conditional operations. Bi-Mamba~\cite{bimamba} and BitNet~\cite{bitnet} also report only packed model size without measured kernel throughput---this is consistent with the current state of the field.

\section{Discussion}

\paragraph{The remaining PPL gap may be training-limited.} The monotonic improvement through 307M tokens (Mann-Kendall $p=0.042$, \S5.4), uniform layer error distribution (\S7), and flat group-size curve (\S5.5) are consistent with the hypothesis that the gap is partially addressable by additional training. The trajectory suggests continued improvement in the 1--10B token range, though we cannot rule out an asymptotic floor imposed by the ternary constraint. Longer QAT with diverse data is a promising next step, potentially complemented by architectural modifications to the quantizer.

\paragraph{Negative results inform future work.} The failure of HG-GSQ (Hessian-guided Gumbel-Softmax, \S A.3) due to the soft-to-hard quantization gap, combined with post-hoc correction failures (\S6), narrows the productive intervention space to: more training tokens, intermediate-layer distillation, and larger model scales where ternary Transformers show quality convergence~\cite{bitnet,paretoq}.

\paragraph{Limitations.} (1) Single architecture and language: all experiments use Mamba-2 1.3B on English (C4); generalization to other SSM variants (Mamba-1, Mamba-3, Jamba) and non-English languages is untested. (2) Only 1.3B scale; the 3B+ quality inflection observed for ternary Transformers remains unverified for SSMs. (3) No custom inference kernel exists---the compression benefit is not yet realized as throughput gain. (4) Sequence length limited to 1024 tokens during training; long-context evaluation is needed.

\section{Conclusion}

Ternary Mamba provides initial evidence that W1.58A16 quantization of SSMs is viable with modest training compute. The method achieves 3.61$\times$ compression with reasonable downstream accuracy (48.1\% vs 55.5\% FP16) in 4 GPU-hours of QAT, identifies zero-ratio collapse as a critical stability constraint, and shows that post-hoc correction strategies designed for Transformers do not readily transfer to recurrent architectures. These findings suggest that future ternary SSM work should prioritize longer QAT schedules, larger model scales, and custom inference kernels over post-hoc output filtering.

\bibliographystyle{plain}
\bibliography{ternary_mamba}

\appendix
\section{Extended Results}

\subsection{Training Hyperparameters}

\begin{table}[h]
\centering
\caption{Training hyperparameters.}
\label{tab:hyperparams}
\begin{tabular}{ll}
\toprule
Hyperparameter & Value \\
\midrule
Base model & \texttt{state-spaces/mamba2-1.3b} \\
Group size & 128 \\
Learning rate & $2.5 \times 10^{-4}$ (cosine, 1k warmup) \\
AdamW $\beta$ & $(0.9, 0.95)$ \\
Weight decay & 0.01 \\
Gradient clip & 1.0 \\
Batch size & $2 \times 1024$ tokens \\
KD $\alpha$ / $T$ & 0.5 / 1.0 \\
Steps & 50,000 \\
Data & C4 (en), local gzip shards \\
\bottomrule
\end{tabular}
\end{table}

\subsection{Parameter Budget}

\begin{table}[h]
\centering
\caption{Parameter budget breakdown.}
\label{tab:params}
\begin{tabular}{lcc}
\toprule
Component & Parameters & Precision \\
\midrule
\texttt{in\_proj} ($\times$48) & 805M & Ternary (1.58-bit) \\
\texttt{out\_proj} ($\times$48) & 403M & Ternary (1.58-bit) \\
embedding + lm\_head & 205M & FP16 \\
SSM params + norms + conv & $\sim$10M & FP16 \\
\midrule
\textbf{Total quantized} & \textbf{1,208M (85.7\%)} & \\
\bottomrule
\end{tabular}
\end{table}

\subsection{Negative Result: HG-GSQ}

Hessian-guided Gumbel-Softmax quantization (targeting top-30\% importance weights with differentiable discrete optimization) degraded PPL by +0.93 to +1.81 vs.\ STE baseline. Root cause: the soft-to-hard gap---the relaxed distribution at $\tau=0.01$ retains $\sim$1\% non-argmax probability mass that the model exploits during training but loses at hard evaluation. STE avoids this by always using hard quantization in the forward pass.

\subsection{Decoupled Quantization Ablations}

Our standard TernaryLinear uses a single scalar per group (absmean) that serves both as the \emph{rounding threshold} (determining which weights map to zero) and the \emph{output magnitude} (scaling non-zero ternary codes). \textbf{Decoupled quantization} separates these into independent parameters: a fixed absmean threshold $\tau_g$ and a learnable magnitude $\alpha_g$, allowing the model to independently control sparsity vs.\ output scale. \textbf{Asymmetric gate thresholds} apply a different rounding threshold ($0.8\times$ absmean) to the gate path (rows 0--4095 of \texttt{in\_proj}, which act as feature selectors) vs.\ the SSM input path (rows 4096--8191), testing whether the gate benefits from lower sparsity.

Results: Learnable magnitude moved only 2.4\% from initialization over 10k steps---the optimizer found no signal to deviate from absmean. Asymmetric gate thresholds mechanically reduced gate zeros (25.7\%$\to$20.9\%) but degraded PPL by +0.12, confirming the model's self-selected uniform sparsity is already optimal. Both interventions fail to improve over simple continued training.

\subsection{FP16 KD Control}

To isolate the KD contribution from the ternary effect: an FP16 model (no quantization) fine-tuned with identical KD setup on WikiText-103 achieves PPL 9.52---below both the off-the-shelf baseline (11.69) and the ternary student (11.25). This confirms ternarization incurs a real +1.73 PPL penalty under controlled conditions, and that exp006's sub-FP16 result is attributable to KD rather than any ternary advantage.

\subsection{Multi-Seed Variance}

\begin{table}[h]
\centering
\caption{Multi-seed replication (3 seeds).}
\label{tab:multiseed}
\begin{tabular}{lccc}
\toprule
Seed & C4 PPL & WT2 PPL & 7-task avg \\
\midrule
42 & 21.66 & 20.45 & 48.3\% \\
137 & 21.68 & 20.29 & 48.1\% \\
2024 & 21.67 & 20.37 & 48.1\% \\
\midrule
\textbf{Mean $\pm$ std} & \textbf{21.67 $\pm$ 0.01} & \textbf{20.37 $\pm$ 0.08} & \textbf{48.1 $\pm$ 0.13pp} \\
\bottomrule
\end{tabular}
\end{table}

Training variance is negligible (C4 PPL std = 0.01), confirming high reproducibility.

\subsection{Per-Layer Quantization Error}

Weight NRMSE is remarkably uniform across all 48 layers (mean 0.5699, CV $<$ 0.005 for both \texttt{in\_proj} and \texttt{out\_proj}). Activation NRMSE decreases from early (1.68) to late layers (0.79), while absolute MSE increases due to growing activation magnitudes. No single layer dominates the total error.

\end{document}